\documentclass[letterpaper, 10 pt, conference]{ieeeconf}  

\IEEEoverridecommandlockouts                              

\newif\ifarxiv
\arxivtrue

\usepackage{color}
\usepackage{amsmath}
\usepackage{amssymb}
\usepackage{graphicx}
\usepackage{comment,xspace}
\usepackage{hyperref}
\usepackage{enumerate}
\usepackage{epstopdf}
\usepackage{subfigure}
\usepackage{cleveref}
\usepackage{tablefootnote}
\usepackage{mathtools}
\usepackage{amsmath,amssymb}
\usepackage{algorithm}
\usepackage{algorithmic}
\usepackage{multirow}
\usepackage{float}
\usepackage{bbm}
\newcommand{\esmargin}[2]{{#1}}
\newcommand{\klmargin}[2]{{#1}}
\usepackage[usenames,dvipsnames,svgnames]{xcolor}

\newcommand{\R}{\mathbb{R}}
\newcommand{\N}{\mathbb{N}}
\newcommand{\E}{\mathbb{E}}
\newcommand{\x}{\mathbf{x}}
\newcommand{\y}{\mathbf{y}}
\newcommand{\z}{\mathbf{z}}
\newcommand*\diff{\mathop{}\!\mathrm{d}}
\DeclareMathOperator*{\argmin}{arg\,min}

\renewcommand{\baselinestretch}{.95}

\title{\LARGE \bf
Multimodal Probabilistic Model-Based Planning \\for Human-Robot Interaction
}

\author{
Edward Schmerling$^{1}$ \quad Karen Leung$^{2}$ \quad Wolf Vollprecht$^{3}$ \quad Marco Pavone$^{2}$%
\thanks{$^{1}$Institute for Computational \& Mathematical \ Engineering, Stanford University, Stanford, CA 94305.
       \{{\tt\small schmrlng@stanford.edu}\}.}%
\thanks{$^{2}$Department of Aeronautics and Astronautics, Stanford University, Stanford, CA, 94305.
       \{{\tt\small karenl7, pavone}\} {\tt \small @stanford.edu}.}%
\thanks{$^{3}$Department of Mechanical Engineering, ETH Zurich, 8092 Zurich, Switzerland.
       \{{\tt\small wolfv@student.ethz.ch}\}.}%
\thanks{This work was supported by the Office of Naval Research (Grant N00014-17-1-2433), by Qualcomm, and by the Toyota Research Institute (``TRI''). This article solely reflects the opinions and conclusions of its authors and not ONR, Qualcomm, TRI or any other Toyota entity.}%
}

\begin{document}

\maketitle
\thispagestyle{empty}
\pagestyle{empty}

\begin{abstract}
This paper presents a method for constructing human-robot interaction policies in settings where multimodality, i.e., the possibility of multiple highly distinct futures, plays a critical role in decision making. 
We are motivated in this work by the example of traffic weaving, e.g., at highway on-ramps/off-ramps, where entering and exiting cars must swap lanes in a short distance---a challenging negotiation even for experienced drivers due to the inherent multimodal uncertainty of who will pass whom.
Our approach is to learn multimodal probability distributions over future human actions from a dataset of human-human exemplars and perform real-time robot policy construction in the resulting environment model through massively parallel sampling of human responses to candidate robot action sequences.
Direct learning of these distributions is made possible by recent advances in the theory of conditional variational autoencoders (CVAEs), whereby we learn action distributions simultaneously conditioned on the present interaction history, as well as candidate future robot actions in order to take into account response dynamics.
We demonstrate the efficacy of this approach with a human-in-the-loop simulation of a traffic weaving scenario.
\end{abstract}

\section{Introduction}\label{sec:intro}

Human behavior is inconsistent across populations, settings, and even different instants, with all other factors equal---addressing this inherent uncertainty is one of the fundamental challenges in human-robot interaction (HRI).
Even when a human's broader intent is known, there are often multiple distinct courses of action they may pursue to accomplish their goals. For example, a driver signaling a lane change may throttle up aggressively to pass in front of a blocking car, or brake to allow the adjacent driver to pass first. 
To an observer, the choice of mode may seem to have a random component, yet also depend on the evolution of the human's surroundings, e.g., whether the adjacent driver begins to yield.
Taking into account the full breadth of possibilities in how a human may respond to a robot's actions is a key component of enabling \klmargin{anticipatory and proactive}{emph?} robot interaction \esmargin{policies}{could maybe use some citations; ONR proposal has some}. With the goal of creating robots that interact intelligently with human counterparts, observing data from human-human interactions provides valuable insight into predicting interaction dynamics \esmargin{\cite{KudererKretzschmarEtAl2012,KueflerMortonEtAl2017,OkamotoBerntorpEtAl2017}}{should be able to find an IRL paper that learns coefficients from human-human interactions}. In particular, a robot may reason about human actions, and corresponding likelihoods, based on how it has seen humans behave in similar settings.
Accordingly, the objective of this paper is to devise a data-driven framework for HRI that leverages learned multimodal human action distributions
in constructing robot action policies.

\begin{figure}[t]
  \centering
  \includegraphics[width=0.5\textwidth]{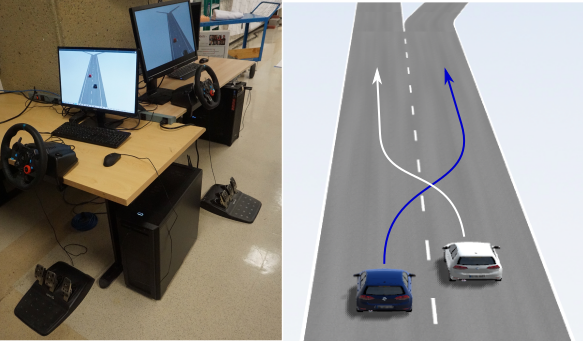}
  \caption{Left: Vires VTD driving simulation setup \cite{VIRESSGH} used to gather a dataset of 1105 pairwise human-human traffic weaving interactions (Right: drivers are required to swap lanes before the highway cutoff; who shall pass whom is ambiguous at the start). We use this dataset to learn a generative model of human driver actions, which is in turn used for exhaustive scoring and selection of candidate robot future action sequences. Our policy is validated on the same simulator for pairwise human-robot traffic weaving interactions.}
  \label{fig:scenario}
\end{figure}

Methods for autonomous decision making under uncertainty may be classified as model-free, whereby human action possibilities and associated likelihoods are implicitly encoded in a robot control policy learned from trial experience, or model-based, whereby a probabilistic understanding of the interaction dynamics is used as the basis for policy construction \cite{Kochenderfer2015}. In this paper we take a model-based approach to pairwise human-robot interaction, seeking to explicitly characterize a possibly multimodal distribution over human actions at each time step conditioned on interaction history as well as future robot action choices. By decoupling action/reaction prediction from policy construction, we
aim to
achieve a degree of transparency in a planner's decision making that is typically unavailable in model-free approaches. Conditioning on history allows a robot to reason about hidden factors like experience, mood, or engagement level that may influence the distribution, and conditioning on the future takes into account response dynamics. We develop our work around a traffic weaving case study (Fig.~\ref{fig:scenario}) for which
we adapt methods from deep neural network-based language and path prediction \cite{SohnLeeEtAl2015,ZhaoZhaoEtAl2017,AlahiGoelEtAl2016,HaEck2017} to learn a Conditional Variational Autoencoder (CVAE) generative model of human driver behavior. We validate this model as the basis for a limited-lookahead autonomous driver action policy, applied in a receding horizon fashion, the behavior of which we explore with human-in-the-loop testing.

We highlight five key considerations that motivate our modeling and policy construction framework. First, for interactive situations, the modeling framework must be capable of predicting human behavior in response to (i.e., conditioned on) candidate robot actions. 
Second, we target decision-making scenarios with characteristic action and response time scales on the order of $\sim$1 second. This is as opposed to \esmargin{higher-level reasoning}{should potentially more explicitly address why we don't consider in this work the possibility that the human just wants to continue straight} over a set of multi-second action sequences or action-generating policies (e.g., whether a driver intends to turn or continue straight at an intersection \cite{GalceranCunninghamEtAl2015}, or intends to initiate a highway lane change \cite{BahramHubmannEtAl2016}). Nor do we attempt to emulate lower-level reactive controllers (e.g., emergency collision avoidance systems \cite{FunkeBrownEtAl2017}) that must operate on the order of milliseconds.
Third, as previously stressed, the uncertainty in human actions on this time scale may be multimodal, corresponding to varied optimal robot action plans. Fourth, we desire a prediction model that is history-dependent, capable of inferring latent features of human behavior. Fifth, although our prediction model is trained ``end-to-end'' from state observations to human action distributions, this work decouples dynamics learning from policy construction to aid interpretability and enable flexibility with respect to robot goals.

\emph{Related Work}:
A major challenge in learning generative probabilistic models for HRI is accounting for the rapid growth in problem size due to the time-series nature of interaction;
state-of-the-art approaches rely on some form of dimensionality reduction to make the problem tractable. One option is to model humans as optimal planners and represent their motivations at each time step as a state/action-dependent cost (equivalently, negative reward) function. Minimizing this function, e.g., by following its gradients to select next actions, may be thought of as a computational proxy for human decision-making processes.
This cost function has previously been expressed as a linear combination of potential functions in a driving context \cite{BahramHubmannEtAl2016,WolfBurdick2008}; Inverse Reinforcement Learning (IRL) is a generalization of this idea whereby a parameterized family of cost functions is fit to a dataset of human state-action trajectories \cite{NgRussell2000,AbbeelNg2004,ZiebartMaasEtAl2008,LevineKoltun2012}. Typically, the cost function is represented as a linear combination of possibly nonlinear features, $c(x,u) = \theta^T \phi(x,u)$, and the weight parameters $\theta$ are fit to minimize a measure of error between the actions that optimize $c$ and the true human actions \cite{NgRussell2000,AbbeelNg2004}.

Maximum entropy IRL interprets this cost function as a probability distribution over human actions, with $p(u) \propto \exp(-c(x,u))$, and in this case the weights $\theta$ are fit according to maximum likelihood \cite{KudererKretzschmarEtAl2012,ZiebartMaasEtAl2008,LevineKoltun2012}. This framework is employed in the context of interactive driving in \cite{SadighSastryEtAl2016} to construct robot policies that avoid being overly defensive by leveraging expected human responses to robot actions. In that work, however, the probabilistic interpretation of $c$ is used only in fitting the human model, not in robot policy construction, where it is assumed that the human selects best responses to robot actions in a Stackelberg game formulation. This analysis yields a unified and tractable framework for prediction and policy construction, but fundamentally represents a unimodal assumption on interaction outcome; we note that this style of reasoning has proven dangerous in the case that critical outcomes go uncaptured \cite{Urmson2016}. Regarding multimodal probabilistic dynamics, it is true that with sufficiently complex and numerous features the cost function $\theta^T \phi(x,u)$ may approximate any log-probability distribution over $u$, conditioned on state $x$, arbitrary well (although we note that IRL is typically applied to learn importance weights for a handful of human-interpretable features). Without some form of state augmentation, however, this formulation is Markovian and incapable of conditioning on interaction history when reasoning about the future.

An additional requirement of cost/motivation-based human modeling methods for use in interactive scenarios is a means to reason about a human's reasoning process. Game theory has been applied to combine human action/reaction inference and robot policy construction \cite{KimLangari2014,NikolaidisNathEtAl2017,LiOylerEtAl2016}. Common choices are Stackelberg formulations whereby human and robot alternate ``turns,'' hierarchical reasoning or ``level-$k$'' approaches whereby agents recursively reason about others reasoning about themselves down to a bounded base case, and equilibria assumption, taking $k\rightarrow\infty$. In this work we approach the human modeling problem phenomenologically, attempting to directly learn the action probability distribution that might arise from such a game formulation.

Computationally tractable human action modeling has also been achieved by grouping actions over multiple time steps and reasoning over a discrete set of template action sequences or action-generating policies \cite{GalceranCunninghamEtAl2015,BahramHubmannEtAl2016,GindeleBrechtelEtAl2010,NikolaidisHsuEtAl2017}.
These methods often tailor means for modeling dependencies between observations, predictions, and latent variables specific to their application that may be used to incorporate information about interaction history. In the context of mutual adaptation for more efficient task execution, \cite{NikolaidisHsuEtAl2017} considers multimodal outcomes and maintains a latent state representing a human's inclination to adapt. The approach of \cite{GalceranCunninghamEtAl2015} for synthesizing autonomous driving policies is very similar in spirit to our work; their prediction model is capable of capturing multimodal human behavior at every time step, dependent on state history and robot future, which they use to score and select from a set of candidate policies for an autonomous vehicle. However, they restrict their treatment of time to changepoint-delineated segments within which the human action distribution takes the form of a Gaussian Mixture Model (GMM). The mean trajectories of these Gaussian components (modes) are predetermined by the choice of a finite set of high-level driving behavior policies, and mixture weight inference takes place over the current time segment.

Inspired by the recent groundbreaking success of deep neural networks in modeling distributions over language sequences \cite{ZhaoZhaoEtAl2017,ChanJaitlyEtAl2015} and geometric paths \cite{AlahiGoelEtAl2016,HaEck2017}, we seek instead to learn a generative model for human behavior that does not decouple time segmentation from probabilistic mode inference, and is capable of learning the equivalent of arbitrary mode policies from data.
Instead of directly fitting a neural network function approximator to log probability, akin to extending maximum entropy IRL with deep-learned features, we use a CVAE \cite{SohnLeeEtAl2015} setup to learn an efficiently sampleable mixture model with terms that represent different driving behaviors over a prediction time horizon. We use recurrent neural networks (RNNs) that maintain a hidden state to iteratively condition each time step's action prediction on the preceding time steps. This opens up the possibility for another level of multimodality within the prediction horizon (e.g., uncertainty in exactly when a human will initiate a predicted braking maneuver), that would otherwise require too many mixture components. An alternate interpretation of why this second level of multimodality is required is to address the case that a mode changepoint, in the language of \cite{GalceranCunninghamEtAl2015}, lies within the fixed prediction horizon. We would also like to briefly mention \cite{OkamotoBerntorpEtAl2017} as a recent nonparametric learning method that compares human driver trajectories to a form of nearest neighbors from an interaction dataset in order to predict future behavior, however, the authors of \cite{OkamotoBerntorpEtAl2017} note that this comparison procedure is difficult to scale at run time for online policy construction.

\emph{Statement of Contributions}:
The primary contribution of this paper is the synthesis and demonstration of a data-driven framework for HRI that, to the best of our knowledge, uniquely meets all five aforementioned conditions for desirable model-based policy construction in rapidly evolving multimodal interaction scenarios. Our human modeling approach, while more data-intensive than some existing alternatives, does not make any assumptions on human preference or indifference between scenario outcome modes, nor does it assume any game-theoretic hierarchy of interaction participant beliefs. The decision framework maintains a degree of interpretability, even as a purely phenomenological data-driven approach, as we are able to visualize through model sampling how the robot anticipates a human might respond to its actions. We demonstrate a massively parallelized robot action sequence selection process that simulates nearly 100\,000 human futures every 0.3 seconds on commodity GPU hardware; this allows us to claim coverage of all possible interaction modes within the planning horizon. This policy is demonstrated in a Model Predictive Control (MPC) fashion in real-time human-robot pairwise traffic weaving simulation. Though the present work is validated in a virtual world for a scenario that, to be frank, may not require such powerful tools as deep neural network-empowered CVAEs or massively parallel GPU simulation for solution, our vision is that as more and more features, e.g., human gestures or verbalizations, beg algorithmic incorporation in more and more complex scenarios, our model-based approach may accommodate them without additional assumptions.

\emph{Notation}: In this paper we use superscripts, e.g., $x^{(t)}$, to denote the values of quantities at discrete time steps, and the notation $x^{(t_1:t_2)} = (x^{(t_1)}, x^{(t_1+1)}, \dots, x^{(t_2)})$ to denote a sequence of values at time steps between $t_1 \leq t_2$ inclusive.

\section{Problem Formulation}
\subsection{Interaction Dynamics}
Let the deterministic, time-invariant, discrete-time state space dynamics of a human and robot be given by
\begin{equation}\label{eqn:dynamics}
x^{(t+1)}_h = f_h(x^{(t)}_h, u^{(t)}_h),\quad
x^{(t+1)}_r = f_r(x^{(t)}_r, u^{(t)}_r)
\end{equation}
respectively, where $x^{(t)}_h \in \R^{d_{x_h}}, x^{(t)}_r \in \R^{d_{x_r}}$ denote the states of the human and robot and $u^{(t)}_h \in \R^{d_{u_h}}, u^{(t)}_r \in \R^{d_{u_r}}$ their chosen control actions at time $t \in \N_{\geq0}$. Let $x^{(t)} = (x^{(t)}_h, x^{(t)}_r)$ and $u^{(t)} = (u^{(t)}_h, u^{(t)}_r)$ denote the joint state and control of the two interaction agents. We consider interactions that end when the joint state first reaches a terminal set $\mathcal{T} \subset \R^{d_{x_h} + d_{x_r}}$ and let $T$ denote the final time step, $x^{(T)} \in \mathcal{T}$. We assume that at each time step $t < T$, the human's next action $u^{(t+1)}_h$ is drawn from a distribution conditioned on the joint interaction history $(x^{(0:t)}, u^{(0:t)})$ and the robot's next action $u^{(t+1)}_r$, that is,
\begin{equation}\label{eqn:human_dist}
U^{(t+1)}_h \sim P\left(x^{(0:t)}, u^{(0:t)}, u^{(t+1)}_r\right)
\end{equation}
is a random variable (capitalized to distiguish from a drawn value $u^{(t+1)}_h$). We suppose additionally that $U^{(t+1)}_h$ is distributed according to a probability density function (pdf) which we write as $p(u^{(t+1)}_h \mid x^{(0:t)}, u^{(0:t)}, u^{(t+1)}_r)$.\footnote{We note that if $U^{(t+1)}_h$ has a discrete component, e.g., a zero acceleration action with positive probability mass, we may add a small amount of white noise to observed values $u$ when fitting distributions for $U^{(t+1)}_h$ to preserve this assumption.} In this work we assume full observability of all past states and actions by both agents. Note that by iteratively propagating \eqref{eqn:dynamics} and sampling \eqref{eqn:human_dist}, a robot may reason about the random variable $U^{(t+1:t+N)}_h$, denoting a human's response sequence to robot actions $u^{(t+1:t+N)}_r$ over a horizon of length $N$.

\subsection{Robot Goal}
We aim to design a limited-lookahead action policy $u^{(t+1:t+N)}_r = \pi_r(x^{(0:t)}, u^{(0:t)})$ for the robot that minimizes expected cost over a fixed horizon of length $N$. We consider a running cost $J(x^{(t)}, u^{(t)}_r, x^{(t+1)})$ which takes the form of a terminal cost $J(x^{(T)}, u^{(T)}_r, x^{(T+1)}) = J_f(x^{(T)})$ should the interaction end before the horizon is reached (with $J(x^{(t)}, u^{(t)}_r, x^{(t+1)}) = 0$ for $t \geq T$). That is, we seek $\pi_r(x^{(0:t)}, u^{(0:t)})$ as an approximate solution to the following minimization problem:
\begin{equation}\label{eqn:robot_cost}
\argmin_{u^{(t+1:t+N)}_r \in \R^{d_{u_r}\times N}} \E\left[\sum_{i=1}^N \gamma^i J(x^{(t+i)}, u^{(t+i)}_r, x^{(t+i+1)}) \right]
\end{equation}
where $\gamma \in [0,1]$ is a discount factor.
Taking $N\rightarrow\infty$ recovers a classical infinite-horizon MDP formulation (although, we note that in this interpretation the state transition distribution is a function of the full state history due to~\eqref{eqn:human_dist}). In practice we take $N = 15$ (with time interval 0.1s) and iteratively solve~\eqref{eqn:robot_cost}, executing only the first action in an MPC fashion. Note that at time $t$ we are solving for the action to take at time step $t+1$, as owing to nonzero computation times we regard the robot action $u^{(t)}_r$ to already be in progress, having been computed at the previous time step.

\subsection{Traffic Weaving Scenario}\label{subsec:traffic_weaving}
Although we have developed our approach generally for pairwise human-robot interactions where the human may be regarded as acting stochastically, we focus in particular on a traffic weaving scenario as depicted in Fig.~\ref{fig:scenario}. 
We learn a human action model and compute a robot policy with the assumption that both agents are signaling an intent to swap lanes.
Let $(s, \tau)$ be the coordinate system for the two-lane highway where $s$ denotes longitudinal distance along the length of the highway (with $0$ at the cutoff point and negative values before) and $\tau$ denotes lateral position between the lanes (with $0$ at the left-most extent of the left lane and negative values to the right). We consider the center of mass of the human-controlled car (``human'') to obey double-integrator dynamics ($u_h = [\ddot{s}_h, \ddot{\tau}_h]$, $x_h = [s_h, \tau_h, \dot{s}_h, \dot{\tau}_h]$ in continuous form). These dynamics may be transformed to various simple car models and map closely to body-frame longitudinal acceleration and steering angle at highway velocities. It is for this latter consideration that we choose a second-order system model, as we believe it most straightforward to fit a generative pdf to inputs on the same order of the true human inputs (throttle and steering command), even if the reward~\eqref{eqn:robot_cost} is a function only of human position or velocity.
We use a similar model for the robot-controlled car (``robot''), but with a triple-integrator in lateral position ($u_r = [\ddot{s}_r, \dddot{\tau}_r], x_r = [s_r, \tau_r, \dot{s}_r, \dot{\tau}_r, \ddot{\tau}_r]$ in continuous form) to ensure a continuous steering command. The robot receives a large penalty for colliding with the human and is motivated to switch lanes before entering the terminal set $\mathcal{T} = \{x=(x_h, x_r) \mid s_r \geq 0\}$ by a running cost proportional to lateral distance to target lane $|\tau_r - \tau^\text{target}_{r}|$ multiplied by an increasing measure of urgency as $s_r$ approaches 0, as well as a reward term that encourages joint states $x$ where the two cars are moving apart longitudinally.
Further details of the cost expression are discussed in Section~\ref{subsec:policy_results}.

A few comments are in order. First, we note that the distributional form~\eqref{eqn:human_dist} admits the possibility that the human has knowledge of the robot's next action before selecting his/her own. This Stackelberg assumption has been employed previously in similar learning contexts \cite{SadighSastryEtAl2016}, but we stress that in the present work this assumption has no bearing on policy construction, where we only sample from $P$. Conditioning on extraneous information should have a negligible effect on learning if $u^{(t+1)}_h$ and $u^{(t+1)}_r$ are truly independent. Second, this formulation is intended to capture factors that may be observed in the prior interaction history $x^{(0:t)}$, e.g., population differences in driving style \cite{LevineKoltun2012}, as well as the response dynamics of the interaction, e.g., game-theoretic behaviors that have previously been modeled explicitly \cite{NikolaidisNathEtAl2017,LiOylerEtAl2016}. Admittedly, approaching this modeling problem phenomenologically, as opposed to devising a more first-principles approach, requires significantly more data to fit. However, with large-scale industrial data collection operations already in place \cite{BojarskiDelTestaEtAl2016}, the ability to utilize large datasets in studying relatively common interaction scenarios, such as the traffic weaving example studied in this paper, is becoming increasingly common.
Third, we note that the robot cost objective sketched above is admittedly rather ad hoc; this is a consequence of requiring the robot to make decisions over a fixed time horizon. We argue that it is difficult to trust interactive prediction models over long horizons $N\gg 0$ in any case, necessitating some sort of long-term cost heuristic to inform short-term actions. Furthermore, our model-based framework can accommodate other cost objectives should they better suit a system designer's preferences, though is not clear what a quantitative measure of quality should be for this traffic weaving scenario. Changing lanes smoothly and courteously, as a human would, does not imply, e.g., that the robot should plan for minimum time or safest possible behavior.

\begin{figure*}[tb]
  \centering
  \includegraphics[width=\textwidth]{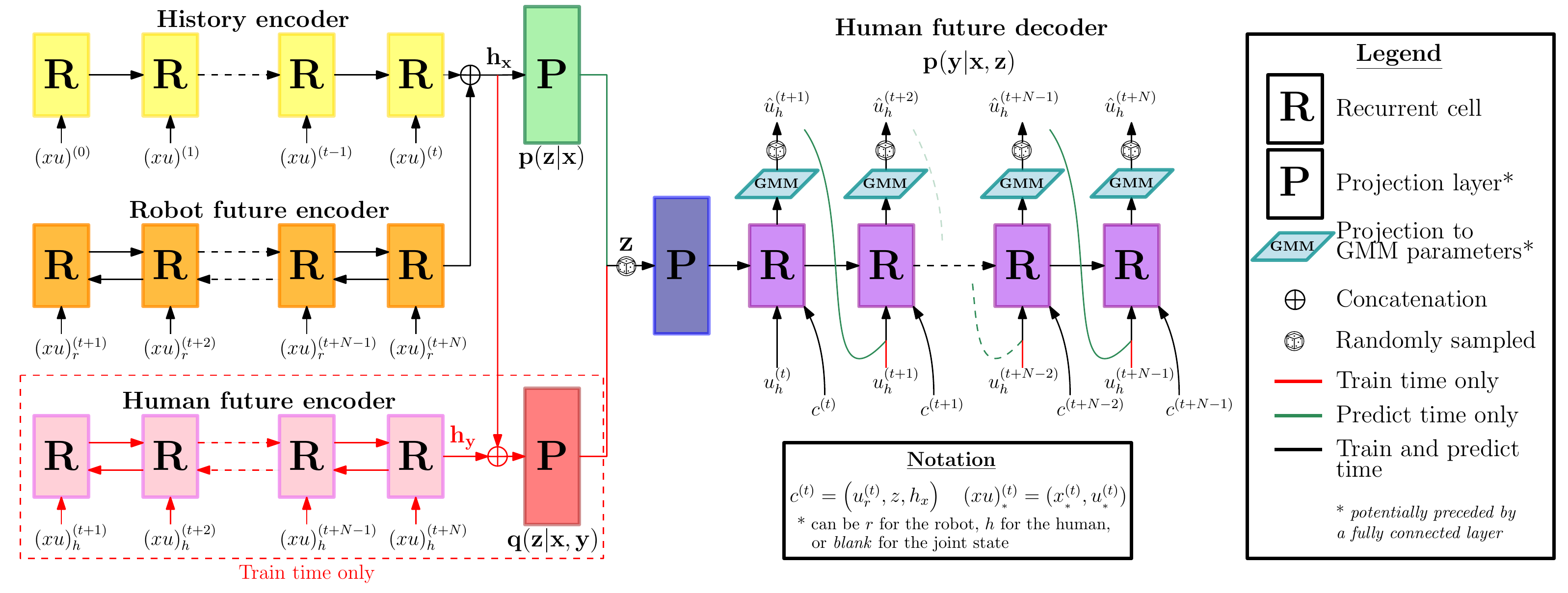}
  \vspace{-0.7cm}
  \caption{CVAE architecture for sequence-to-sequence generative modeling of future human actions $\y = u_h^{(t+1:t+N)}$ conditioned on joint interaction history $(x^{(0:t)}, u^{(0:t)})$ and candidate robot future actions $u^{(t+1:t+N)}_r$ (together, $\x$). The random variable $\z$ is a latent mixture component index.}
  \vspace{-0.5cm}
  \label{fig:cvae}
\end{figure*}

\section{Human Agent Modeling}\label{sec:learning}

Inspired by recent breakthroughs in sequence-to-sequence learning for language modeling \cite{ChanJaitlyEtAl2015,BowmanVilnisEtAl2015} and geometric path generation \cite{HaEck2017,ZhangYinEtAl2017}, as well as unsupervised end-to-end learning of complex conditional distributions \cite{SohnLeeEtAl2015,ZhaoZhaoEtAl2017}, we design our learning framework as a neural network Conditional Variational Autoencoder (CVAE) with recurrent subcomponents to manage the time-series nature of the interaction. Considering a fixed prediction time step $t$, let $\x = (x^{(0:t)}, u^{(0:t)}, u^{(t+1:t+N)}_r)$ be the conditioning variable (joint interaction history + candidate robot future) and $\y = u^{(t+1:t+N)}_h$ be the prediction output (human future). That is, we seek to learn a pdf $p(\y | \x)$. The CVAE framework introduces a latent variable $\z$ so that $p(\y | \x) = \int_\z p(\y | \x, \z)p(\z | \x) \diff\z$; the intent of introducing such a $\z$ is to model inherent structure in the interaction which may both improve learning performance as well as provide a foothold for interpretation of the results \cite{SohnLeeEtAl2015}. We model $p(\y | \x, \z)$ and $p(\z | \x)$ using neural networks with parameters fit to maximize likelihood of a dataset of observed interactions $\mathcal{D} = \{(\x_i, \y_i)\}_i$.\footnote{In practice, log-likelihood $\log p(\y_i | \x_i) = \log \int_\z p(\y_i | \x_i, \z)p(\z) \diff\z$ is stochastically optimized by importance sampling $\z$ according to a neural network approximation $q(\z | \x_i, \y_i)$ of the true posterior $p(\z | \x_i, \y_i)$, and the evidence-based lower bound (ELBO) on log-likelihood is maximized instead. For a detailed discussion see \cite{Doersch2016}.} Our CVAE architecture, depicted in Fig.~\ref{fig:cvae}, is similar to existing sequence-to-sequence variational autoencoder architectures \cite{ZhaoZhaoEtAl2017,HaEck2017} including the use of RNN encoders/decoders; we highlight here a few key design choices that we feel are essential to our model's success.

\subsection{CVAE Architecture}

We choose a discrete latent space such that $\z$ is a categorical random variable with distribution factored over $N_\z$ independent elements, $p(\z = (z_1,\dots z_{N_\z})) = \prod_{i=1}^{N_\z}p(z_i)$ where each element has $K_{\z}$ categories, resulting in a latent space dimension of size $K_{\z}^{N_{\z}}$. Thus $p(\y | \x) = \sum_
\z p(\y | \x, \z)p(\z | \x)$ may be thought of as a mixture model with components corresponding to each discrete instantiation of $\z$. The learned distributions in our work have $\z$ values that manifest as different human response modes, e.g., accelerating/decelerating or driving straight/turning over the next $N$ time steps, as illustrated in Section~\ref{subsec:distribution_results}. Modeling this multimodal human response behavior is a core focus of this work; the benefits of using a discrete latent space to learn multimodal distributions has been previously studied in \cite{JangGuEtAl2017,MaddisonMnihEtAl2017,MoerlandBroekensEtAl2017}.

In our model, the discrete latent variable $\z$ has the responsibility of representing high-level behavior modes, while a second level of multimodality within each such high-level behavior is facilitated by an autoregressive RNN sequence decoder (light purple cells, Fig.~\ref{fig:cvae}). The RNN maintains a hidden state to allow for drawing each future human action conditioned on the actions drawn at previous future times:
\begin{equation}\label{eqn:RNN_breakdown}
p(\y | \x, \z) = \prod_{i=1}^N p(\y^{(i)} | \x, \z, \y^{(1:i-1)})
\end{equation}
where we use the notation $\y^{(i)} = u_h^{(t+i)}$, the $i$-th future human action after the current time step $t$. A GMM output layer of each RNN cell with $M_\text{GMM}$ components in human action space provides the basis for learning arbitrary distributions for each $p(\y^{(i)} | \x, \z, \y^{(1:i-1)})$ (light blue, Fig.~\ref{fig:cvae}). This combined structure is designed to account for variances in human trajectories for the same latent behavior $\z$. For example, an action sequence of braking at the fourth time step instead of at the third time step would likely need to belong in a different component of, e.g., a non-recurrent Gaussian Mixture Model, causing an undue combinatorial explosion in the required number of distinct $\z$ values.

We use long short-term memory (LSTM) RNN cells \cite{HochreiterSchmidhuber1997}, and in addition to typical model training techniques (e.g., recurrent dropout regularization \cite{SemeniutaBarth2016} and hyperparameter annealing), we employ the method introduced in \cite{BengioVinyalsEtAl2015} to prevent prediction errors from severely cascading into the future. During train time in the decoder (right side, Fig.~\ref{fig:cvae}), with a rate of $10\%$ we use the predicted value $\hat{u}_h^{(t)}$ as the input into the next cell, otherwise we use the true value from the training data. That is, instead of learning individual terms of~\eqref{eqn:RNN_breakdown}, we occasionally learn them jointly.
Similar to \cite{HaEck2017} we augment the human action inputs for the autoregressive decoder RNN with an additional context vector $c^{(t)}$ composed of the robot action at that time step $u_r^{(t)}$, the latent variable $\z$ and the output from the encoder $h_x$. We do this to more explicitly mimic the form of the expression $p(\y^{(i)} | \x, \z, \y^{(1:i-1)})$.

\section{Robot Policy Construction}
We apply an exhaustive approach to optimizing problem \eqref{eqn:robot_cost} over robot action sequences where at each time step the robot is allowed to take one of a finite set of actions (in particular, we group the $N=15$ future time steps into five 3-step windows over which the robot takes 1 of 8 possible actions, see Section~\ref{subsec:policy_results} for details). We stress that we are only discretizing the robot action space for policy computation efficiency; the human prediction model is still computed over a continuous action space. While Monte Carlo Tree Search (MCTS) methods have seen successful application in similar problems with continuous state/action spaces \cite{CoueetouxHoockEtAl2011}, due to the massively parallel GPU implementation of modern neural network frameworks \cite{Abadi2015} we find it is more expedient to simply evaluate the expected cost of taking all action sequences (or a significant fraction of them) for sufficiently short horizons $N$. We take a two step approach, approximately evaluating all action sequences with a low number of samples (human response futures) each, and then reevaluating the most promising action sequences with a much larger number of samples to pick the best one (see Section~\ref{subsec:policy_results}). We note that gradient-based methods could prove useful for further action sequence refinement, but for scenarios characterized by multimodal outcomes some form of broader search must be applied lest optimization end in a local minimum.

\section{Traffic Weaving Case Study}
The human-human traffic weaving dataset and source code for all results in this section, including all network architecture details and hyperparameters, are available at \url{https://github.com/StanfordASL/TrafficWeavingCVAE}. All simulation and computation was done on a computer running Ubuntu 16.04 equipped with a 3.6GHz octocore AMD Ryzen 1800X CPU and an NVIDIA GeForce GTX 1080 Ti GPU.
\subsection{Data Collection}
\begin{figure}[tb]
  \centering
  \includegraphics[width=0.5\textwidth]{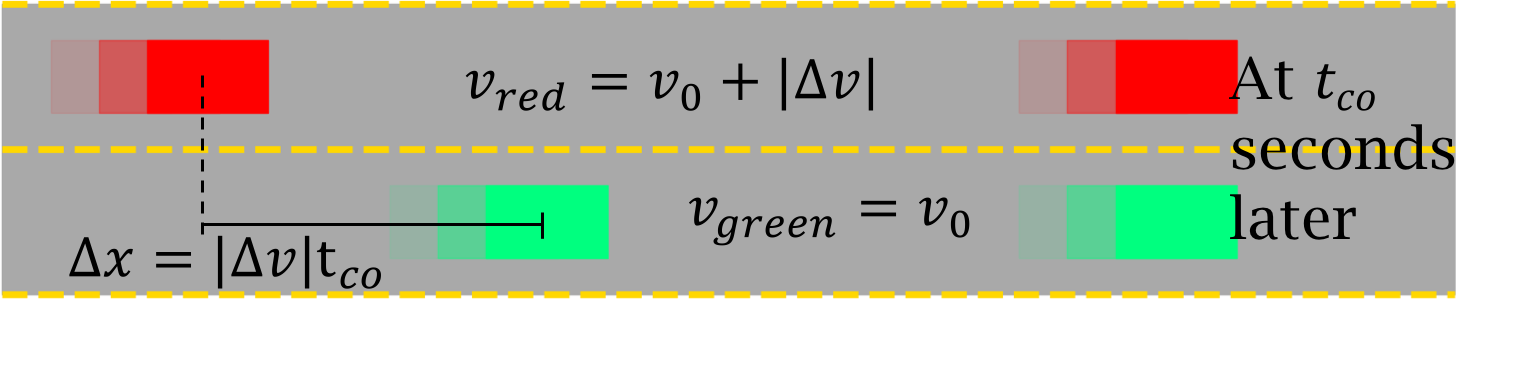}
  \vspace{-0.8cm}
  \caption{Traffic weaving scenario initial conditions: with zero control actions by both cars, their longitudinal coordinates will cross over in $t_{co}$ seconds.}
  \label{fig:tco_descrip}
\end{figure}

State and action trajectories of two humans navigating a traffic weaving scenario were collected using a driving simulator \cite{VIRESSGH} shown in Fig.~\ref{fig:scenario} (left). 1105 human-human driving interaction trials were recorded over 19 different pairs of people. The drivers had to swap lanes with each other (without verbal communication) within 135 meters of straight road. Each human trajectory in each trial exhibits interaction behavior to be learned, effectively doubling the data set.
Furthermore, since we are conditioning on interaction history and partner future, cumulatively speaking we have roughly 35 histories per trial: each trial is approximately five seconds long, equating to $T \approx 50$ with 0.1s time steps, and taking the prediction horizon of $N=15$ into account.
Hence in total, our dataset contains roughly 77\,000 
$\x = (x^{(0:t)}, u^{(0:t)}, u^{(t+1:t+N)}_r)$ to $\y = u^{(t+1:t+N)}_h$
exemplars.
We note that owing to the interactive nature of this scenario we elected to collect our own dataset using the simulator rather than use existing real-world data (e.g., \cite{ColyarHalkias2007,KotserubaRasouliEtAl2016,RasouliKotserubaEtAl2017}). Fitting the parameters of our model requires a high volume of traffic weaving interaction exemplars which these open datasets do not encompass, but which we believe a targeted industrial effort might easily procure. Each scenario begins with initial conditions (IC) drawn randomly as follows: car 1 (left or right lane) starts with speed $v_1 = 29$m/s (65mph) and car 2 at a speed difference of $\Delta v = v_1 - v_2 \in  \lbrace 0, \pm 2 \rbrace$ m/s ($\pm 4.5$ mph). The faster car starts a distance $|\Delta v|t_{co}$ behind the other car where $t_{co}\in \lbrace 1, 2, 3 \rbrace$ represents a ``crossover time'' when the cars would be side-by-side if neither accelerates or decelerates, as shown in Fig.~\ref{fig:tco_descrip}. The ICs were designed to make ambiguous which car should pass in front of the other, to encourage multimodality in action sequences and outcomes that may occur. This ambiguity is indicated by Table~\ref{tab:initial_condition_statistics} where as $t_{co}$ increases, the car moving faster but starting further behind, which usually passes in front of the other car, becomes less likely to cut in front.
Reckless and irresponsible ``video game'' driving was discouraged by having a speedometer displayed on screen with engine hum sound feedback to reflect current speed and a high-pitched alert sound when speed exceeded 38m/s (85mph).

\begin{table}[tb]
  \caption[caption]{Fraction of human-human trials where\\ car starting in left lane passes in front.}
  \label{tab:initial_condition_statistics}
  \centering
  \begin{tabular}{|c|c|c|c|}
  \hline
  &  \multicolumn{3}{|c|}{Relative speed $\Delta v = v_\text{left} - v_\text{right}$}\\
  \hline
  \bf $t_{co}$ & -2m/s & 0m/s & 2m/s \\
  \hline 
  1s &  7.2 \% & \multirow{3}{*}{38.2\%} & 85.7\% \\
  2s & 13.1 \% & & 76.2\%\\
  3s & 23.6 \% & & 66.7\%\\
  \hline
  \end{tabular}
\end{table}

\subsection{Generative Model for Human Driver Actions}\label{subsec:distribution_results}

\begin{figure*}[tb]
  \centering
  \includegraphics[width= \textwidth]{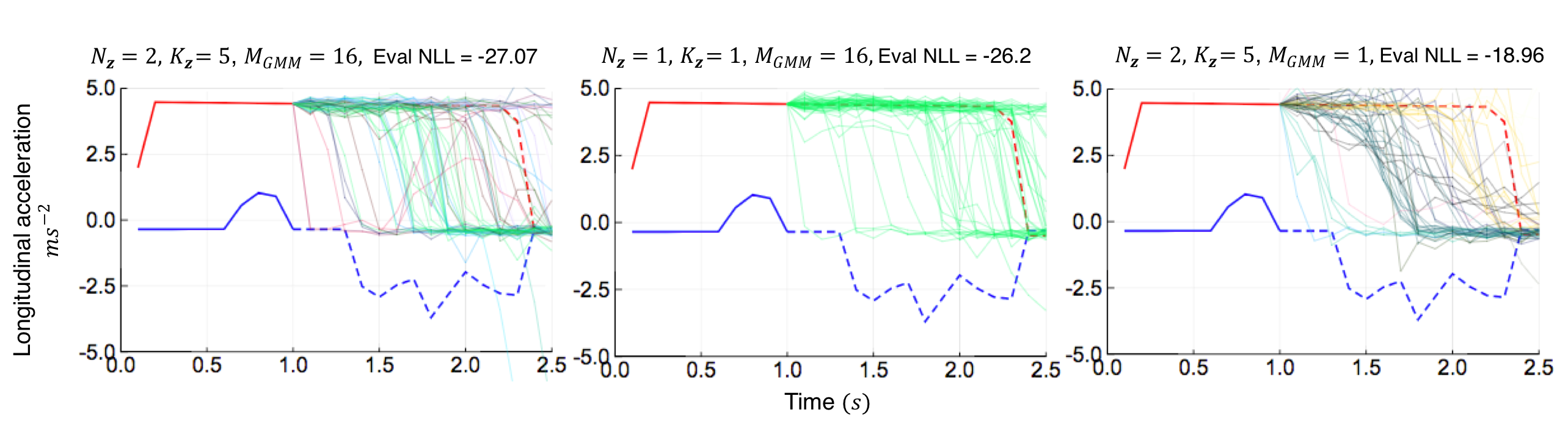}
  \vspace{-0.9cm}
  \caption{Predictions (50 samples) from three different models of human future longitudinal acceleration actions conditioned on the available interaction history (solid lines) and future robot actions (dashed blue line). The dashed red line represents the true future action sequence taken by the human. The line color of the prediction corresponds to different $\z$ values, if applicable. Eval NLL denotes negative log-likelihood prediction loss over a fixed validation set of interaction trials (lower is better).
  Not Pictured: basic LSTM prediction model (no multimodality, with $N_{\z} = K_{\z} = M_\text{GMM} = 1$, Eval NLL = -9.70).}
  \vspace{-0.3cm}
  \label{fig:prediction_panel}
\end{figure*}

We implemented the neural network architecture discussed in Section~\ref{sec:learning} in Tensorflow \cite{Abadi2015} and fit human driver action distributions over a 1.5s time horizon conditioned on all available history up to prediction time, as well as the next 1.5s of the other driver's trajectory (i.e., a robot's candidate plan). Fig.~\ref{fig:prediction_panel} illustrates why we chose to design these action distributions as mixtures indexed by a discrete latent variable $\z$, where the component distributions are a combination of recurrent hidden state propagation and GMM sampling. All three models capture the same broad prediction: eventually, the human will cease accelerating. Comparing the left plot to the middle plot, the CVAE addition of a latent $\z$ manifests as modes predicting cessation on a few different time scales; this unsupervised clustering aids interpretability and slightly improves the performance metric of validation set negative log-likelihood. Comparing the right plot, of essentially a mixture of basic LSTM models, to the rest, we see that neglecting multimodality on the time step to time step scale prevents the prediction of sharp behavior (i.e., a human's foot quickly lifted off of the throttle).

\subsection{Robot Policy Construction}\label{subsec:policy_results}

\begin{figure}[h]
  \centering
  \includegraphics[width=0.5\textwidth]{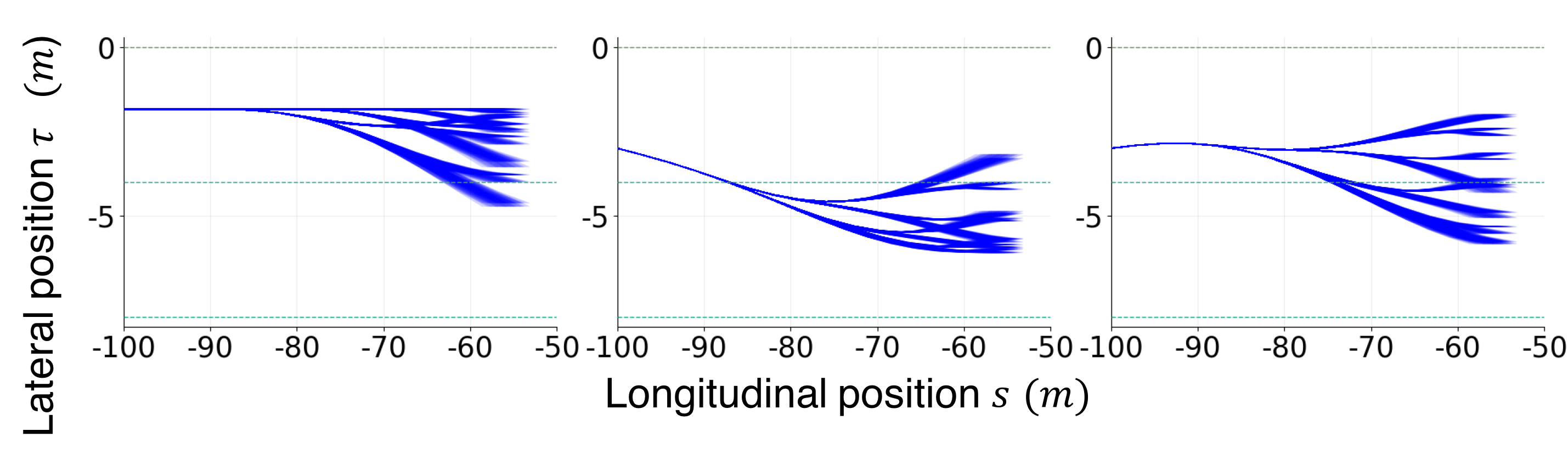}
  \vspace{-0.7cm}
  \caption{4096 candidate robot action sequences scored each planning loop.}
  \vspace{-0.3cm}
  \label{fig:4096}
\end{figure}

We consider a discrete set of robot candidate futures over the 1.5s prediction horizon. We target a replanning rate of 0.3s and break the prediction horizon up into five 0.3s fixed action windows within which the robot may choose one of four longitudinal actions, $\ddot{s}_r \in \{0, 4, -3, -6\}\text{m/s}^2$, and one of two lateral actions, moving towards either the left lane or the right lane. Specifically, the robot control $\dddot{\tau}_r$ is selected as the first 0.3s of the optimal control for the two point boundary value problem steering from $(\tau_0, \dot{\tau}_0, \ddot{\tau}_0)$, at the start of the window, to $(\tau_\text{target}, 0, 0)$, at some free final time $t_f$ after the start of the window, with cost objective $\int_0^{t_f} 1 + \dddot{\tau}_r^2/1000\diff t$. In total the robot has 8 possible actions per time window; given that actions in the first window are assumed fixed from the previous planning iteration this results in $8^4 = 4096$ possible action sequences the robot may select from. These 4096 sequences are visualized in $(s,\tau)$ coordinates for a few initial robot states in Fig.~\ref{fig:4096}.

\begin{figure*}[tb]
  \centering
  \includegraphics[width=\textwidth]{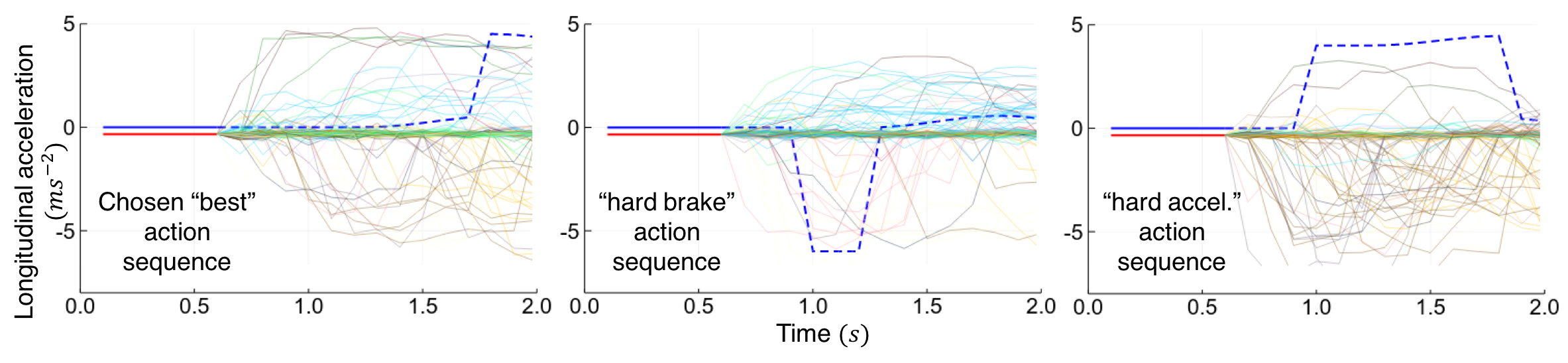}
  \vspace{-0.9cm}
  \caption{Example of considerations the robot makes when choosing the best candidate action sequence. Left: the best-scoring sequence is the robot first waiting then planning to eventually accelerate since it has time to wait and see if the human will accelerate or slow down.
  Center: a candidate action sequence showing a braking maneuver which predicts the human likely to accelerate (in order pass in front) in response. Right: a candidate robot future action sequence showing a hard acceleration maneuver which predicts the human likely to decelerate (in order to let the robot pass) in response. However, these sequences (center and right) incur a higher control cost penalty $J_a$ than waiting as in the action sequence the exhaustive policy search chooses.}
  \vspace{-0.3cm}
  \label{fig:prediction_policy_synthesis}
\end{figure*}

The robot's running cost consists of four terms $J(x^{(t+i)}, u^{(t+i)}_r, x^{(t+i+1)}) = J_c + J_a + J_l + J_d$ corresponding to collision avoidance, control effort, lane change incentive, and longitudinal disambiguation incentive defined as:
\begin{align*}
J_c &= 1000\cdot\mathbf{1}_{\{|\Delta s| < 8 \wedge |\Delta\tau| < 2\}}\cdot(9.25 - \sqrt{\Delta s^2 + \Delta\tau^2})\\
J_a &= \ddot{s}_r^2\qquad J_l = -500\cdot\min(1.5 + s_r/150, 1)\cdot|\tau_r - \tau_\text{goal}|\\
J_d &= -100\cdot\min(\max(\Delta s \Delta \dot{s}, 0), 1)
\end{align*}
where we have omitted a time index $(t+i)$ from all state/action quantities, and $\Delta s = s_r - s_h$, $\Delta \tau = \tau_r - \tau_h$, $\Delta \dot{s} = \dot{s}_r - \dot{s}_h$. Briefly, $J_c$ is a radial penalty past a near-collision threshold, $J_a$ is a quadratic control cost on longitudinal acceleration (lateral motion is free), $J_l$ is proportional to distance from target lane centerline ($\tau_\text{goal}$) with an urgency weight that increases as the robot nears the end of the road at $s = 0$, and $J_d$ incentivizes reaching states where $\Delta s$ is the same sign as $\Delta \dot{s}$, i.e., the two cars are moving apart from each other longitudinally. We use a discount factor $\gamma = 0.9$.

We draw 16 samples of human responses to each of the 4096 candidate robot action sequences and average their respective costs to approximate the expectation in problem~\eqref{eqn:robot_cost}. We select the top 32 sequences by this metric for further analysis, scoring 1024 sampled human trajectories each to gain a confident estimate (up to the model fidelity) of the true cost of pursuing that robot action sequence. This two-stage sampling and scoring process for our prediction model takes $\sim$0.25s parallelized on a single GTX 1080 Ti, simulating nearly 100\,000 human responses in total. The robot action sequence with the lowest expected cost from the second stage is selected for enactment over the next action window. In particular, the second action window of the sequence is propagated next (as the first was already being propagated as the policy computation was running) and becomes the fixed first action window of the next search iteration.

\subsection{Human-in-the-Loop Example Trials}\label{subsec:results_discussion}

\begin{figure}[tb]
  \centering
  \includegraphics[width=0.5\textwidth]{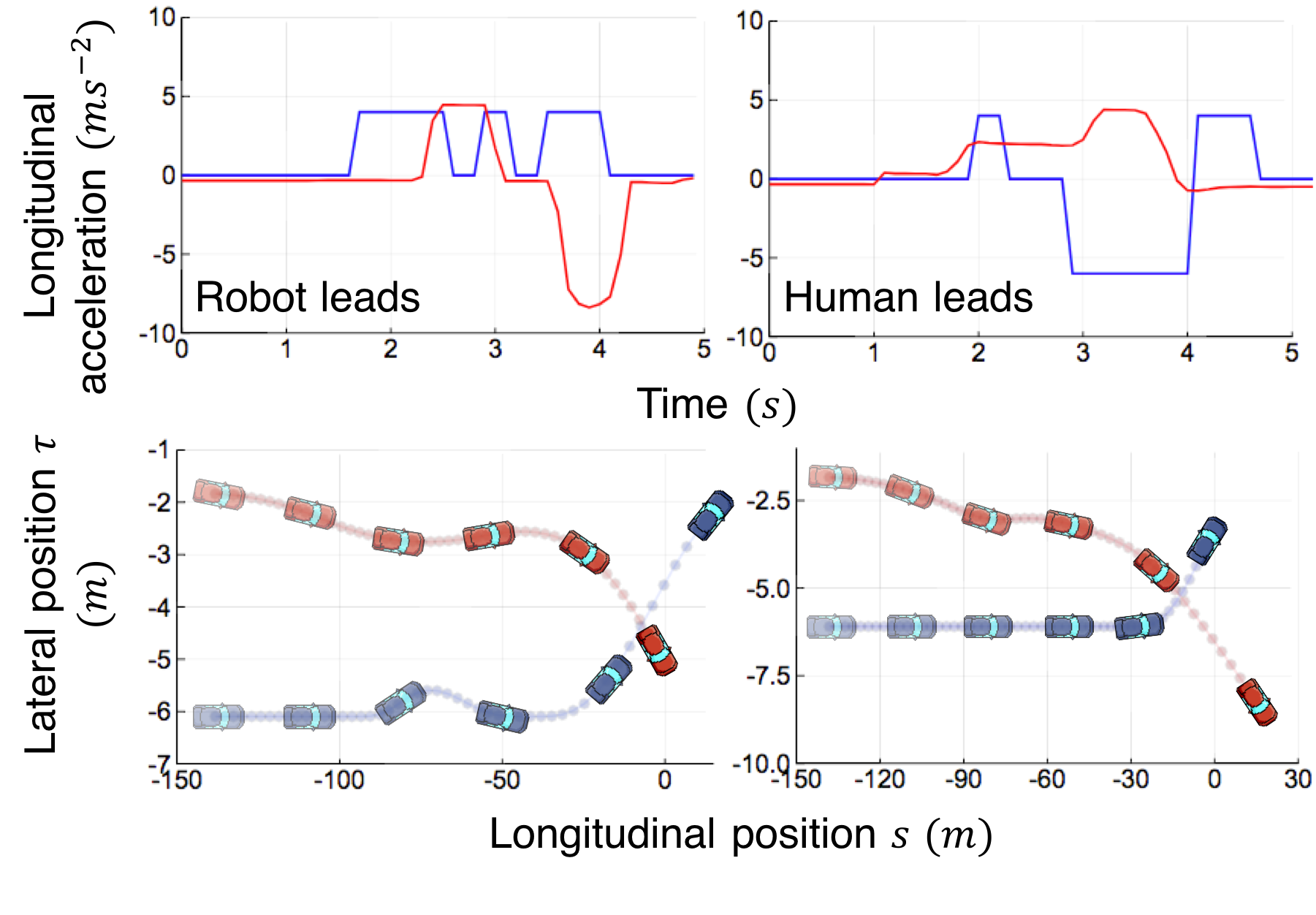}
  \vspace{-0.9cm}
  \caption{Interaction action sequences and $(s,\tau)$-trajectories of the robot (blue) and a human driver (red) who tries to cut in front. Top Left: robot chooses to accelerate, causing the human to brake to change lanes near the end of the road.  Top Right: both the human and robot accelerate at first, but the human is more persistent and cuts off the robot. Bottom: Lateral and longitudinal positions of the human and robot car, showcasing successful autonomous human-in-the-loop traffic-weaving maneuvers.}
  \vspace{-0.3cm}
  \label{fig:hitl_reactions}
\end{figure}
This robot policy was integrated with the simulator, Fig.~\ref{fig:scenario}, enabling real-time human-in-the-loop validation of the robot action sequences it selects. As in the human-human input dataset, the intent of both parties is to switch lanes. Though, as noted in Section~\ref{subsec:traffic_weaving}, the desired behavior is qualitative and near-impossible to reliably quantify (especially human-in-the-loop), we highlight here some interesting emergent behaviors in robot reasoning. Fig.~\ref{fig:prediction_policy_synthesis} illustrates an example of the robot's decision making at a single time step early in the interaction. The robot is aware of multiple possible actions the human might take, and even aware of how to elicit specific interaction modes, but chooses to wait in accordance with the sequence that minimizes its expected cost~\eqref{eqn:robot_cost}. Fig.~\ref{fig:hitl_reactions} illustrates two examples where both the human and robot tried to be proactive in cutting in front of the other. When the human does not apply control actions early on, the robot nudges towards the lane divider expecting the human to yield, showcased in the left figures. The trial in the right figures starts out much the same, but the human's continued acceleration causes the robot to change its mind and brake to let the human pass. We note that the robot's cost function contains no explicit collaboration term, meaning it is essentially fending for itself while reasoning about relative likelihoods from its human action model. This is not unlike many drivers on the road today, but we note that our framework accommodates adjusting the robot cost but keeping the human model the same---friendlier behavior may be achieved through adding or changing cost terms.

\section{Conclusions}
We have presented a robot policy construction framework for HRI that takes as input (1) a dataset of human-human interaction trials, in order to learn an explicit, sampleable representation of human response behavior, and (2) a cost function defined over a planning horizon, so that the desired robot behavior within the interaction model may be achieved through exhaustive action sequence evaluation applied in an MPC fashion. This framework makes no assumptions on human motivations nor does it rely on reasoning methods or features designed specifically for the traffic weaving scenario (other than the robot cost objective); it learns relative likelihoods of future human actions and responses at each time step from the raw state and action dataset. As such the robot is essentially blind to what it has not seen in the data---this framework is designed for probabilistic reasoning over relatively short time horizons in nominal operating conditions, but an important next step is to integrate it with lower-level emergency collision avoidance routines and higher-level inference algorithms, e.g., what if the human is not explicit in signaling its intent to change lanes? Another promising avenue of future research is to incorporate existing work in reasoning hierarchies into the human model learning architecture. Finally, our ultimate vision is to get this framework off the simulator and onto the road, where we may begin to benefit over alternative frameworks from our ability to incorporate additional learning features at will.

\bibliographystyle{IEEEtran-short}
{
\bibliography{../../../../bib/main,../../../../bib/ASL_papers}
}

\end{document}